\documentclass{article}

\usepackage{PRIMEarxiv}

\usepackage[utf8]{inputenc} 
\usepackage[T1]{fontenc}    
\usepackage{hyperref}       
\usepackage{url}            
\usepackage{booktabs}       
\usepackage{amsfonts}       
\usepackage{nicefrac}       
\usepackage{microtype}      
\usepackage{lipsum}
\usepackage{fancyhdr}       
\usepackage{graphicx}       
\usepackage{amsmath}
\usepackage{algorithm2e}
\usepackage{multirow}
\usepackage{float}
\graphicspath{{media/}}     

\title{Isolating authorship from content with semantic embeddings and contrastive learning
\thanks{\textit{\underline{Citation}}: 
\textbf{Authors. Title. Pages.... DOI:000000/11111.}} 
}

\author{
  Javier Huertas-Tato, Adrián Girón-Jiménez, Alejandro Martín, David Camacho \\
  Dept. of computer systems \\
  Universidad Politécnica de Madrid \\
  Madrid\\
  \texttt{\{javier.huertas.tato, adrian.giron, alejandro.martin, david.camacho\}@upm.es} \\
}

\begin{document}
\maketitle

\begin{abstract}
Authorship has entangled style and content inside. Authors frequently write about the same topics in the same style, so when different authors write about the exact same topic the easiest way out to distinguish them is by understanding the nuances of their style. Modern neural models for authorship can pick up these features using contrastive learning, however, some amount of content leakage is always present. Our aim is to reduce the inevitable impact and correlation between content and authorship. We present a technique to use contrastive learning (InfoNCE) with additional hard negatives synthetically created using a semantic similarity model. This disentanglement technique aims to distance the content embedding space from the style embedding space, leading to embeddings more informed by style. We demonstrate the performance with ablations on two different datasets and compare them on out-of-domain challenges. Improvements are clearly shown on challenging evaluations on prolific authors with up to a 10\% increase in accuracy when the settings are particularly hard. Trials on challenges also demonstrate the preservation of zero-shot capabilities of this method as fine tuning.
\end{abstract}


\section{Introduction}
Automated authorship analysis textual content has been subject of scientific interest study in recent years. Understanding the nuances of authorship expression is a computationally challenging task that has broad range of application from social media moderation to text forensics. Computational methods have been developed towards authorship (and style) understanding in recent years with success. An interesting task in this domain is Authorship Attribution (AA): given an unknown text, discover the author, relying on a database of already known reference documents. This task has been useful for training authorship-centric models in the past years, as it is a suitable objective for semi-supervised and supervised contrastive learning approaches.

When analysing authorship it is to be noted that an author usually writes about the same topic. A fantasy author frequently writes about fantasy, while a political twitter user will write about politics most of the time. If a machine learning model is trained to understand authorship, it is reasonable to expect that it will automatically correlate the writing topic as something characteristic of an author. In other words, an author favourite topic contaminates the model, offering a shortcut for the neural network to abuse. Problems arise when two authors write about the same topic where the model can easily be deceived by the spurious correlation inserted by the writing topic. We call this phenomenon Style-Content Entanglement (SCE), and it is an undesirable property of neural networks trained towards AA and authorship style understanding.

\begin{figure}
    \centering
    \includegraphics[width=0.95\textwidth]{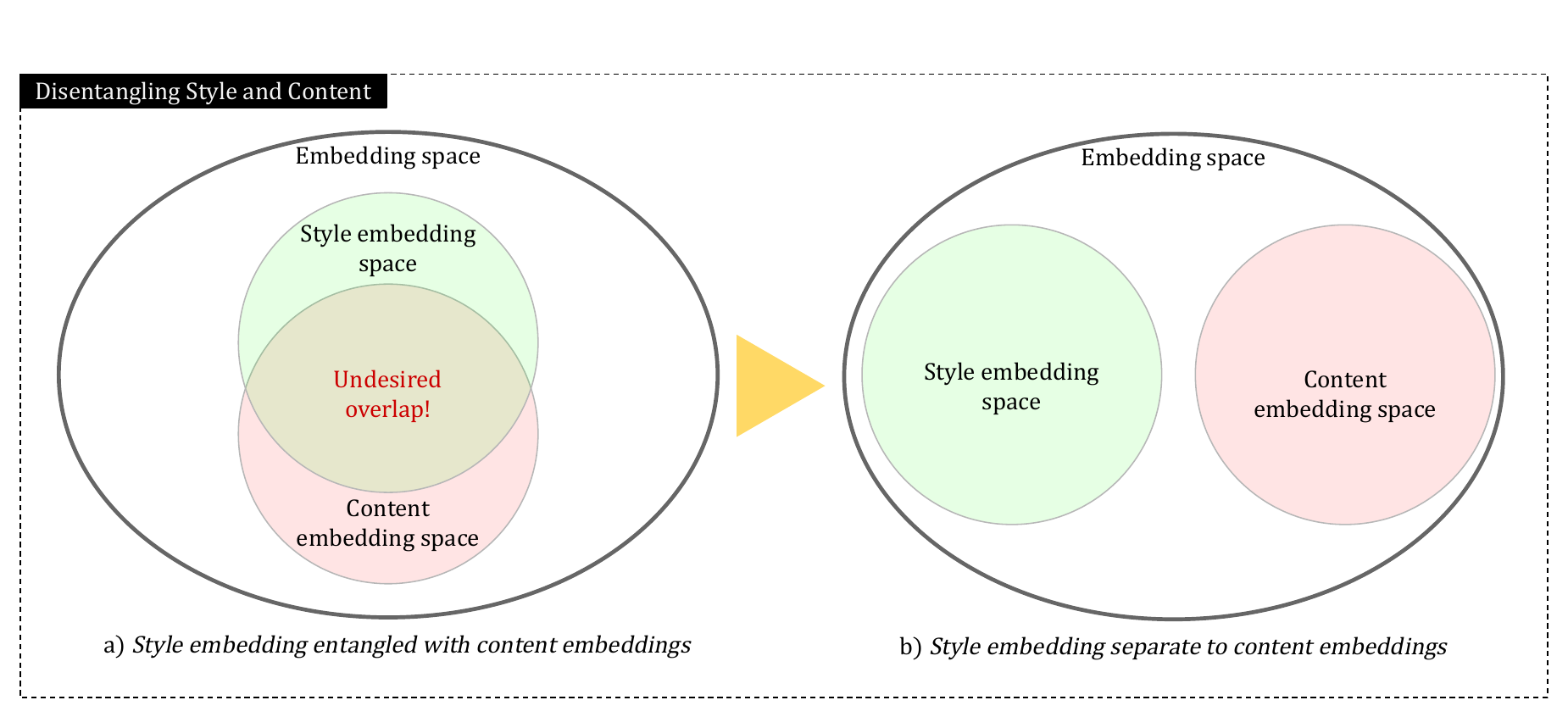}
    \caption{Disentangling embedding spaces for style and content}
    \label{fig:abstract}
\end{figure}

The objective of this article is to disentangle effectively style and content. In the possible hyperspace for an N-dimensional embedding there exist a smaller subspace after the model has been trained towards style. This also happens when a model is trained to understand content, as it is the case of Masked Language Modeling (MLM) or Causal Language Modeling (CLM); where the model is trained and generates embeddings within a subspace. We hypothesize that these subspaces themselves are entangled, leading to inaccuracies in the embeddings themselves. Thus, to build an effective method for disentanglement we can use a method to produce content embeddings and show them to the training objective as negative examples, leading to a separation between the style embedding space and the content embedding space. This key idea is represented in Fig.~\ref{fig:abstract}.

Transformers represent the state-of-the-art in language understanding task and style analysis in particular. It is possible to use pre-trained models to achieve our objectives. Style-oriented models have been successfully applied to AA problems~\cite{huang2024authorship}, while content-oriented models dominate the state of the art on text tasks. Using a pre-trained style model we can warm-start our method from an already functioning style embedding space, while using a pre-trained language model allows us to have a reference content embedding space.

In this paper we present a neural network method disentanglement method that extends InfoNCE~\cite{oord2018representation} loss to include seamlessly deceptive examples from another space. This method reports increased accuracy on AA tasks when authors write about several topics, overcoming some barriers of the SCE issue. Our main contributions are the following:
\begin{itemize}
    \item Modified InfoNCE loss capable to effectively disentangle two embedding spaces
    \item Method to improve authorship models by disentangling style and content
    \item Ablation study that demonstrates that the disentangling method contributes to higher accuracies
\end{itemize}

The article shows the related work in this area (Sec.~\ref{sec:related}) followed by our presented method (Sec.~\ref{sec:method}), the specific configuration of our experiments as well as the description of our data (Sec.~\ref{sec:methodology}). Finally, section~\ref{sec:results} shows our empirical results and discussion paired with section~\ref{sec:conclusions}.

\section{Related work}
\label{sec:related}
The rapid expansion of online media has transformed various fields, particularly Natural Language Processing, where an unprecedented volume of text is generated daily. This vast collection of written content opens up new opportunities to utilize advanced transformer models for authorship analysis using sources like blogs and digital libraries. In this section, we explore authorship as it pertains to our research, reviewing methods that enable a quantitative examination of stylistic and authorship nuances.

Writing inherently reflects the individual who authors it—their beliefs, knowledge, habits, and more influence any extended text they create. Based on this premise, we define authorship as a unique set of textual features identifiable with a specific individual. Authorship attribution, for instance, links these stylistic features to their source. 

\subsection{Stylometric-Based Prediction for Authorship Attribution}

In the field of authorship attribution, there is a general consensus on considering authorship as the "form" of a text, where predictive models are employed over stylometric features~\cite{ouni2023survey, tyoStateArtAuthorship2022}. These features can be categorized into three primary types: (1) lexical features~\cite{palomino-garibayRandomForestApproach, daneshvarGenderIdentificationTwitter}, such as word frequency or character n-grams; (2) syntactic features~\cite{abbasiWriteprintsStylometricApproach2008}, including punctuation and part-of-speech (POS) tags; and (3) structural features~\cite{ashrafCrossGenreAuthorProfile, fernquistFourFeatureTypes}, such as vocabulary richness and sentence length. These methods are grounded in the idea that each author's unique writing style reveals identifying characteristics, such as punctuation habits, the use of certain expressions, or average sentence length. Consequently, approaches that track word occurrences can effectively capture these stylistic elements. 

Common text vectorization methods, such as Term Frequency-Inverse Document Frequency (TF-IDF)~\cite{sammutTFIDF2010} at the word level or n-grams at the character level, have been widely adopted in authorship attribution. Sari et al.~\cite{sariContinuousNgramRepresentations2017} used continuous n-gram representations, while Coyotl-Morales et al.~\cite{coyotl-moralesAuthorshipAttributionUsing2006} combined functional and content features to model the frequency of sub-phrases within documents by specific authors. In a related study, Jacobo et al.~\cite{jacoboAuthorshipVerification} used two vectorization methods: a term-document matrix capturing term frequency in the text and prediction by pair matching (PPM), a technique that predicts the next symbol based on preceding ones. Muttenthaler et al.~\cite{muttenthalerAuthorshipAttributionFanFictional} employed an ensemble model approach by combining predictions across different n-gram levels.

However, these methods may underperform in broader contexts or when authors are discussing similar topics~\cite{hu2024ContrastiveDisentanglement}. Our approach aims to address these challenges by seeking a representation that captures the author's style independently of content, providing a more generalizable solution across varied domains.

\subsection{Relying in Semantics-Biased Pre-training}

With the recent advancements of Transformer models in natural language processing (NLP)~\cite{vaswani2017attention}, several studies have investigated how the linguistic knowledge acquired during pre-training might contribute to authorship attribution (AA) tasks. Transformer-based models like BERT~\cite{devlin2019bert}, primarily trained on large corpora using objectives such as Masked Language Modeling (MLM) for semantic understanding and Next Sentence Prediction (NSP) for syntactic coherence, are often hypothesized to retain some stylistic knowledge implicitly. However, the focus of these pre-training tasks remains on achieving a generalized understanding of language, primarily at the semantic and syntactic levels, raising questions about their effectiveness in author-specific subtasks such as AA.

Fabien et al.~\cite{fabienBertAABERTFinetuning} explored this idea by fine-tuning BERT specifically for AA, combining it with stylometric features like word frequency and mixed features like n-grams. Their approach integrates BERT classification probabilities with additional features using logistic regression. In prior research, we introduced PART model~\cite{huertas2022part}, built over the hypothesis that pre-trained models such as RoBERTa~\cite{liu2019roberta} might capture stylistic nuances with a proper fine-tuning. By employing contrastive learning, PART maximizes the similarity between representations of texts by the same author and minimizes it for different authors, seeking to capture inherent style characteristics.

However, pre-training approaches generally prioritize content over style, which can complicate the model’s ability to discern nuanced stylistic differences between authors covering similar topics. Our work diverges from this trend by focusing on isolating style-specific features, reducing the influence of content-focused pre-training and thereby enhancing the model's capability to distinguish authorship in content-similar texts. Even style focused works like PART also pick up on content-based features due to authors publishing frequently about similar topics.

\subsection{Disentangling style and content}

Traditional methods frequently struggle under conditions where multiple authors discuss similar topics or when stylistic nuances are subtle. Recent approaches have aimed to address these challenges by separating content and style representations or by using contrastive learning to improve the robustness of authorship predictors.

Romanov et al.~\cite{romanov2019AdversarialDecomposition} introduced ADNet, a GAN-based model~\cite{goodfellowGenerativeAdversarialNets2014a} for adversarial decomposition, which separates text into distinct vectors for form and meaning. The architecture uses an encoder to extract these latent vectors, while a generator reconstructs the sentence by combining them. ADNet’s discriminator enforces adversarial loss on the meaning vector to exclude form information, while a motivator encourages retention of style in the form vector. Extending this idea, Hu et al.~\cite{hu2024ContrastiveDisentanglement} addressed a gap in adversarial attribution (AA) methods, particularly under topic-shift conditions that lead to style-content conflation. Their model, ContrastDistAA, trains a style encoder with supervised contrastive loss and introduces mutual information minimization, successfully disentangling content from style.

Wegmann et al.~\cite{wegmann2022same} advanced authorship verification (AV) by employing a contrastive framework that compares an anchor utterance with two others—one by the same author. This design allows control over the utterances’ contextual origin (e.g., same conversation, domain, or random), enhancing AV accuracy by leveraging context-sensitive cues. Sawatphol et al.~\cite{sawatphol2022TopicRegularizedAuthorship} developed Authorship Representation Regularization (ARR), which generates author representations less influenced by topic-specific content. ARR utilizes an encoder and a TF-IDF-based topic similarity model, creating a topic-regularized probability score that guides the distillation into a final author classification.

In previous work, we introduced the STAR model~\cite{huertas2023understanding}, using supervised contrastive learning to bring texts by the same author closer in vector space while pushing apart those from different authors. This revealed content biases, particularly in cases where multiple authors addressed similar topics. This study takes a content-agnostic approach, producing separate content and style representations. Our model’s training is structured as a multi-objective task, with the main goal of minimizing the distance between style representations of texts authored by the same individual, allowing for more accurate authorship attribution independent of content similarities.

\section{Authorship and content}
\label{sec:method}
Successful authorship attribution (AA) neural methods use contrastive learning (CL) objectives. Methods using CL try to maximize a distance metric between positive and negative examples, while minimizing this metric between positives. The end result of the CL process is a model that embeds some inputs into an N-dimensional space. In our domain, positive examples are represented by text documents that belong to a single author, while negative examples represent any other text document by a different author. This approach is both simple and effective at building style embedding spaces, as demonstrated by previous work~\cite{wegmann2022same,huertas2022part,huertas2023understanding}.

Under close examination, we observe that authors usually repeat the topic of their writing. When we assign texts of the same author as the positive examples, we frequently indirectly assign texts of the same topic as positive examples, which teaches the model to produce similar embeddings when the topics are also similar. Although expected from authorship, this is an undesirable side-effect of CL because content is not unique to an author, leading the model to find high similarity between authors writing about the same topic: a spurious correlation.

\begin{figure}
    \centering
    \includegraphics[width=0.5\textwidth]{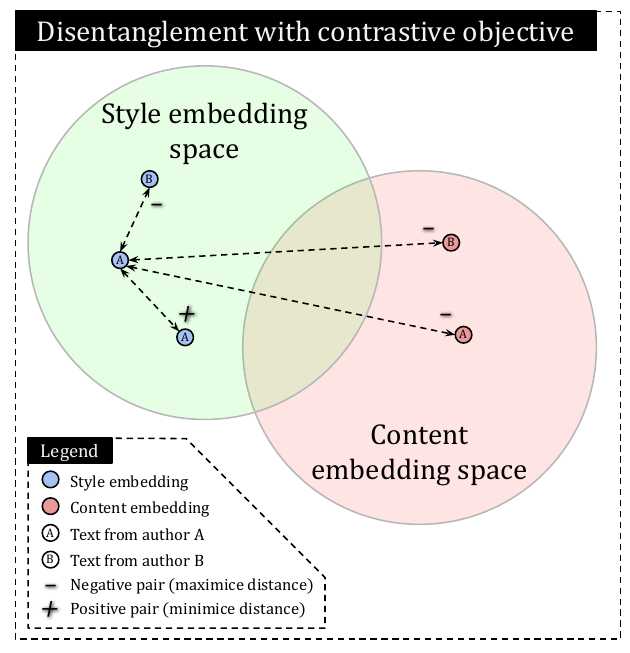}
    \caption{Contrastive learning objective for disentanglement}
    \label{fig:idea}
\end{figure}

To counteract this, we modify our objective function to differentiate style from content entirely. We maximize similarity when texts belong to the same author, while minimizing similarity from texts belonging to different authors, while simultaneously maximizing distance between embeddings generated by the style model from embeddings generated by the content model. This disentanglement with a contrastive objective is presented in Fig.~\ref{fig:idea}. 

A contrastive objective designed with the previous approach would build a style space separate from the static content embedding space generated by an auxiliary model. All points in the style subspace should be dissimilar to all points in the content subspace. This implies that the disentanglement process is strictly reliant on the content embedding model's capabilities, assuming that the subspace generated by the embeddings of this model does not contain any trace of style, which is obviously untrue. Therefore, it is likely that this style embedding will retain some amount of topic due to imperfections in the content embedding space of the auxiliary content model. Our focus is on using a high-quality surrogate model.

\subsection{Problem statement}
Let $D_{A} = \{D_{A, 1}, D_{A, 2}, D_{A, 3}, ..., D_{A, N}\}$ and $D_{B} = \{D_{B, 1}, D_{B, 2}, D_{B, 3}, ..., D_{B, N}\}$ be two sets of $N$ documents, where $D_{A, i}$ and $D_{B, i}$ represents a pair of documents corresponding of the $i$-th author. No author is repeated inside each document set. Let $s(\cdot)$ be a trainable model, that receives a document as input and outputs an embedding as in $s(D_{A,i}) = S_{A,i}$ where $S_{A,i}$ is the style embedding for the document set $A$ and author $i$. Likewise there is $c(\cdot)$ that is a frozen model for content that similarly produces $c(D_{A,i}) = C_{A,i}$ where $C_{A,i}$ is a content embedding. Following this defined approach we want to minimize a distance function (cosine distance namely $S_{cos}(\cdot, \cdot)$). There are four optimizations in our method either minimizing distance or maximizing it:
\begin{itemize}
    \item Minimize style distance between similar authors: $S_{cos}(S_{A,i}, S_{B,i}) \forall i \in \{1,..., N\}$
    \item Maximize style distance between different authors $S_{cos}(S_{A,i}, S_{B,j}) \forall i \in \{1,..., N\}, j \in \{1,..., N\} : i\neq j$
    \item Maximize style-content self-distance $S_{cos}(S_{A,i}, C_{A,j}) \forall i \in \{1,..., N\}, j \in \{1,..., N\}$
    \item Maximize style-content distance to foreign set $S_{cos}(S_{A,i}, C_{B,j}) \forall i \in \{1,..., N\}, j \in \{1,..., N\}$
\end{itemize}

For this method we implement an adapted version of InfoNCE~\cite{oord2018representation}. We first need to build a reference set as defined in Eq.~\ref{eq:reference}:
\begin{equation}
    \label{eq:reference}
    R(D_A, D_B) = s(D_B) \mathbin\Vert c(D_A) \mathbin\Vert c(D_B)
\end{equation}
We define access to the reference set as $R(D_A, D_B)_k$ where $k \in \{1,...,3N\}$. Later we define the probability of the positive pair among all pairs. The main difference with InfoNCE is the inclusion of additional negative pairs from the described reference set $R(D_A, D_B)$. We describe the likelihood of a positive pair to be maximized in Eq.~\ref{eq:probability}

\begin{equation}
    \label{eq:probability}
    P(R(D_A, D_B)_j \text{ is positive } | D_{A,i}) = \frac{\text{exp }(\frac{S_{cos}(D_{A,i}), R(D_A, D_B)_j}{\tau})}{\sum^{3N}_{k=1} \text{exp }(\frac{S_{cos}(D_{A,i}), R(D_A, D_B)_k}{\tau})
    }
\end{equation}

where $\tau$ is a trainable temperature parameter. The aforementioned expression can be used to compute the Cross-Entropy Loss ${L}_{CE}$ of a single document $D_{A,i}$ in terms of Eq.\ref{eq:cross} \begin{equation}
    \label{eq:cross}
    \mathcal{L}_{CE}(D_{A,i}, R(D_A, D_B)) = -\sum^{3N}_{j=1}
    \text{log }P(R(D_A, D_B)_j \text{ is positive } | D_{A,i})
\end{equation}

To compute the loss over all documents we reduce using the average of all documents, as in Eq.\ref{eq:final_cross}:
\begin{equation}
    \label{eq:final_cross}
    \mathcal{L}(D_A, D_B) = \frac{1}{N}\sum^N_{i=1} \mathcal{L}_{CE}(D_{A,i}, R(D_A, D_B))
\end{equation}

\begin{figure}
    \centering
    \includegraphics[width=0.95\textwidth]{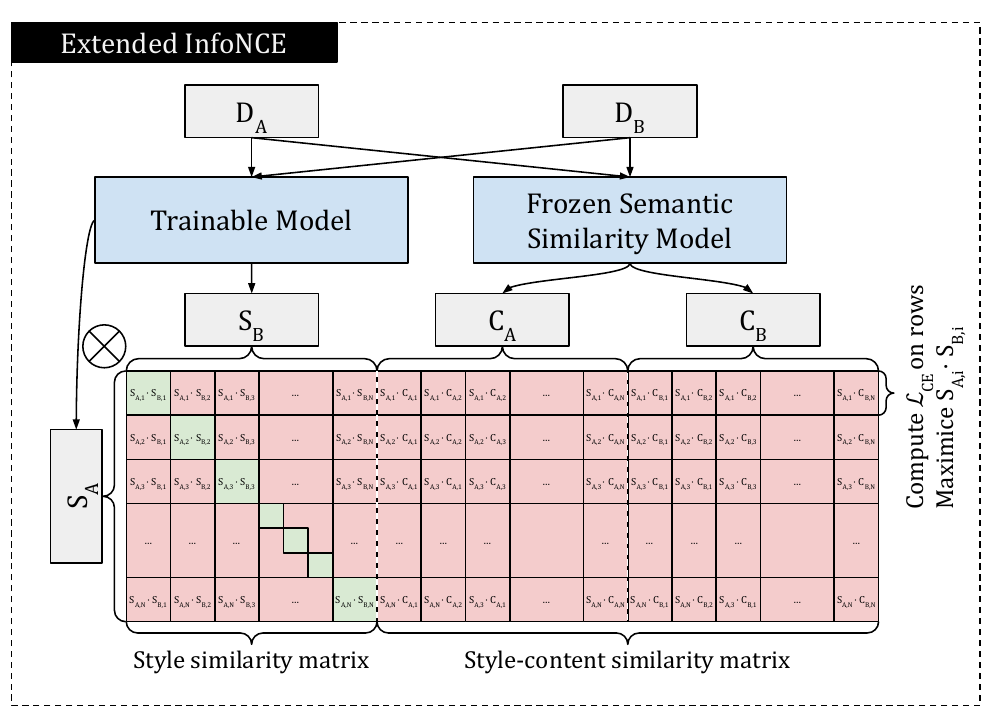}
    \caption{Contrastive learning objective for disentanglement}
    \label{fig:objective}
\end{figure}

A visual overview of this process is described in Figure~\ref{fig:objective}. For simplicity, we compute the cosine similarity matrices of each embedding set pair and then build a matrix concatenating all three, computing $\mathcal{L}_CE$ for each row of the full similarity matrix. This process is finally implemented with Algorithm~\ref{alg:pseudo}.

\RestyleAlgo{ruled}
\SetKwComment{Comment}{/* }{ */}
\begin{algorithm}
\caption{Custon InfoNCE Python pseudocode}\label{alg:pseudo}
\KwData{$D_A, D_B$}
$N = ||D_A||$\Comment*[r]{cardinality of document set}
$S_A = \text{style\_model}(D_A)$\Comment*[r]{output=(batch\_size,embedding\_dim)}
$S_B = \text{style\_model}(D_B)$\;
$C_A = \text{content\_model}(D_A)$\;
$C_B = \text{content\_model}(D_B)$\;
$sim_1 = \text{cosine\_distance }(S_A, S_B)$\Comment*[r]{output=(batch\_size, batch\_size)}
$sim_2 = \text{cosine\_distance }(S_A, D_A)$\;
$sim_3 = \text{cosine\_distance }(S_A, D_B)$\;
$sim_f = \text{cat}(\{sim_1, sim_2, sim_3\}, \text{dim}=-1)$\;
$labels = \text{range}(0, N)$ \Comment*[r]{for code we begin at 0 to N-1}
$sim_f = sim_f - \text{max}(sim_f, \text{dim}=1).detach()$ \Comment*[r]{normalization step for stability}
$\text{\textbf{return}} \text{ cross\_entropy}(sim_f, labels)$

\end{algorithm}

\section{Methodology}
\label{sec:methodology}
To verify the correct performance of our model we design the following experimentation. We use two pre-trained model, the first towards warm-starting the style embedding space to reduce training cost, the second toward obtaining a high quality semantic embedding space in reasonable computational time. We employ two datasets to test our methods, as well as standard authorship attributions competitions. We also provide the hyper parameters sets for our experimentation as well as technical details on our setup.

\subsection{Model choices}
We require a model architecture for the style part of the algorithm, and a pre-trained language model for the frozen content encoder. First, we have chosen to warm-start our method with a pre-trained style model to reduce the training time of our algorithm. This includes special consideration, as we include the original frozen style encoder as part of our experimentation, as well as a fine-tuned style encoder with only regular InfoNCE. For the frozen content encoder we require a robust model of similar output embedding dimension to our style encoder. We rely on encoder-only models with specific focus on semantic similarity. Here we outline the results of our previous decisions:

\textbf{STAR}~\cite{huertas2023understanding}\footnote{Model Checkpoint: \href{https://huggingface.co/AIDA-UPM/star}{AIDA-UPM/star}}: In our previous work we pre-trained a model for authorship understanding. It reports high performance on several attribution benchmarks. It was trained using a Supervised Contrastive Learning objective without content disentanglement. If our reasoning is sound, the detangling will improve performance in scenarios where authors explore heterogeneous topics, which was an identified problem of the aforementioned model.

\textbf{UAE}~\cite{li2023angle}\footnote{Model Checkpoint: \href{https://huggingface.co/WhereIsAI/UAE-Large-V1}{WhereIsAI/UAE-Large-V1}}: This model performs semantic similarity at a state-of-the-art level against large language models with only a fraction of the parameters. This model was optimized towards semantic similarity using AnglE loss. 

Furthermore, we use the following strategy to validate our method against three different baselines. First, we use RoBERTa to ensure that the trained models require more than simple semantic understanding. STAR is our next baseline, ensuring that zero-shot evaluation is not enough to attribute on some datasets. Finally the last baseline is the fine-tuned version of STAR without the usage of semantic embeddings, a simple fine-tuning using InfoNCE with the same parameters, ensuring that adapting to the task itself is not enough to increase performance beyond content. At least for the trials on the fine-tuned datasets, the detangling procedure should increase performance above all three baselines simultaneously.

\subsection{Data}
We test our methods with several datasets. The first is the PAN competitions, we use these to benchmark our model against an array of top-performing models in the same category, usually custom-tailored to each task which makes it more competitive. Also we require data to detangle and validate said detanglement, where we will explore the performance in two domains, a medium-scale dataset and a larger dataset with richer categories. 

\textbf{Pre-processing:} For training purposes we extract chunks of 512 tokens using the respective tokenizer of each model. Most posts and literary works have many more than 512 tokens, thus we design a randomized truncation policy: we select a random point in the token sequence to extract a full 512 sequence of tokens. We need at least 2 documents per author (one for document set $D_A$ and document set $D_B$) thus any author with less than 2 documents is discarded. Works with unknown authorship are discarded too. For style preservation we avoid removing punctuation, data augmentation or other techniques that may alter the original style of a document; as such we perform minimal preprocessing aside from the stated tokenization and data cleaning.

The data has been split in three partition for training, validation and testing with 80\% of authors at train and 10\% at validation and test. We split on authors instead of documents to examine the attribution capabilities at unseen writing styles.

We provide descriptions of the data and their contribution to our experiments:

\textbf{Validation}: To validate our models we perform authorship attribution on the datasets. To validate we randomly choose a document from an author and try to attribute each document to their authors using the remaining documents. For stability in our metrics we perform 10 trials with random selection. Each dataset contains different metadata which we will use independently to obtain different metrics from the authorship attribution tasks, we detail each under the \textit{validation strategy} section of each dataset.

\textbf{Blog dataset}~\cite{JonathanSchler2023Oct}: This datasets contains texts from more than 19.000 bloggers written before 2004. They contain several posts per blogger with an average of 35 posts. They are categorized with labels relating to gender, age and, most importantly, topic. There are less than 30 different occupations in total, so the dataset is really narrow topic-wise, with a majority writing about an unknown occupation or being students. \textit{Validation strategy}: We perform three experiments, first a global attribution results across the entire dataset, attribution results with authors constrained by their publishing topic, and finally we observe to extra metrics: the accuracy given the reference topic, and the proportion of misses when the topic is identical. We aim to maximize the accuracy given the reference topic while minimising the percentage of misses due to similar topic, to guarantee that the model does not misclassify examples due to similarities in the topic.

\textbf{Fanfiction.net archive crawl}~\cite{BibEntry2014Mar}: This dataset is large in terms of texts and authors containing >100GB of fan-fiction works with >70k different topics (the subject of the fan-fiction) and more than >1.3M different authors with a total >6M posts (An average of 5 works by author). Considering an average of 8k words per work, the dataset is one of the largest available for this task. \textit{Validation strategy}: We conducted two experiments. First, we assessed the overall effectiveness of our method. Second, we focused on a specific subset of authors with diverse writing styles. This second evaluation is crucial because style attribution models often struggle to accurately attribute styles across authors who share similar topics but publish on different subjects.

\textbf{PAN attribution challenge}~\cite{juola2013overview, stamatatos2014overview, kestemont2018overview, kestemont2019overview}: We use all available PAN authorship attribution challenges to test our fine-tuned models. These challenges are varied in domain from PAN2013 emails to PAN2019 fan-fiction. It is to be noted that there may be overlap with the Fanfiction.net archive crawl on the PAN2019 dataset. These challenges serve towards the purpose of testing our fine-tuned models for detanglement and check if they retain their out-of-domain capabilities. \textit{Validation strategy}: In this set validation is straightforward, where we simply run the attribution challenges as designed.

\subsection{Technical setup}
Here we describe technicalities about our article regarding hardware, hyper-parameters, code repository and so on.
\textbf{Hyper-parameters}
In total we perform 4 training passes for each algorithm (one for simple fine-tuning and our detangling fine tuning). All training have been performed using the hyper-parameters as presented in Table~\ref{tab:hparams}. These hyper-parameters have been extracted from our previous work, epochs have been adjusted for dataset size (in terms of number of authors) and learning rate has been adjusted for optimal performance from a set of \{0.001, 0.002, 0.005, 0.01\}. 

\begin{table}[H]
\caption{Hyper parameter settings}
\centering
\label{tab:hparams}
\begin{tabular}{@{}lcc@{}}
\toprule
\textbf{Parameter}          & \textbf{Fanfiction}     & \textbf{Blog}    \\ \midrule
\textit{Batch Size}         & \multicolumn{2}{c}{2048}                   \\
\textit{Embedding dim.}     & \multicolumn{2}{c}{1024}                   \\
\textit{Dropout}            & \multicolumn{2}{c}{0.1}                    \\
\textit{Optimizer}          & \multicolumn{2}{c}{Adam w/ weight decay}   \\
\textit{Learning Rate}      & 0.001                   & 0.005            \\
\textit{Scheduler}          & \multicolumn{2}{c}{Warmup w/ linear decay} \\
\textit{Warmup Steps}       & \multicolumn{2}{c}{0.06}                   \\
\textit{Epochs}             & 25                      & 250              \\ \bottomrule
\end{tabular}
\end{table}

\textbf{Technical details}
Training and validation has been performed on a Titan V GPU over the course of a month given the described hyperparameters. We have used the PyTorch \footnote{Pytorch: \href{https://pytorch.org/}{pytorch.org}} backend with the Lightning API \footnote{Pytorch Lightning: \href{https://lightning.ai/docs/pytorch/stable/}{lightning.ai/docs/pytorch/stable/}} to perform and log the experiments. Our code is available at \href{https://lightning.ai/docs/pytorch/stable/}{lightning.ai/docs/pytorch/stable/}

\section{Results}
\label{sec:results}
We present our experiments in the following section, evaluating our test partitions and the challenges. We end the results with a discussion over the obtained results. 

\subsection{Blogs}
The summary of our experiments over the blog dataset are reflected in Table~\ref{tab:blog-sum}. As we demonstrate the fine-tuning with detanglement outperforms with 4\% difference in Accuracy and Macro f1-score the simple finetune. As we hypothesized, on average, fitting to the task is not enough to achieve better performance. Although fine-tuning itself benefits the performance in attribution, it is much more beneficial to detangle with an increase of 8\% instead of a 4\% increase, which doubles the absolute improvement for the task. Also, accuracy is very low when simply doing semantic similarity with RoBERTa, resulting in the lowest score available.

\begin{table}[H]
\centering
\label{tab:blog-sum}
\caption{Attribution results across the entire blog author set. Repeated 10 times.}
\begin{tabular}{lrrr}
\toprule
    & \textbf{Accuracy} & \textbf{F1 Score} \\
\midrule
\textit{RoBERTa}                 & 0.2445 & 0.4534 \\
\textit{STAR}                    & 0.3463 & 0.4716 \\
\textit{Fine-tune (simple) }     & 0.3928 & 0.5196 \\
\textit{Fine-tune (detangle) }   & \textbf{0.4469} & \textbf{0.5588} \\
\bottomrule
\end{tabular}
\end{table}

\begin{figure}[H]
    \centering
    \includegraphics[width=0.8\linewidth]{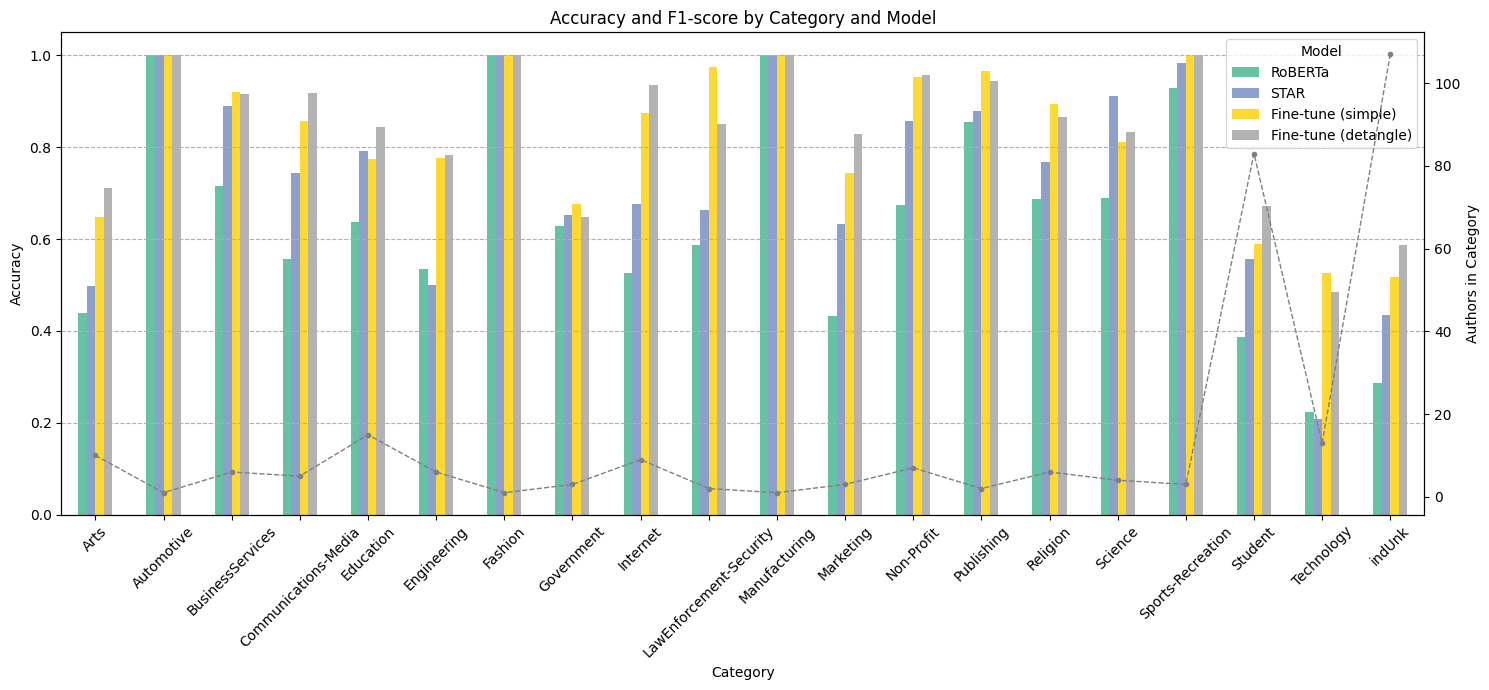}
    \includegraphics[width=0.8\linewidth]{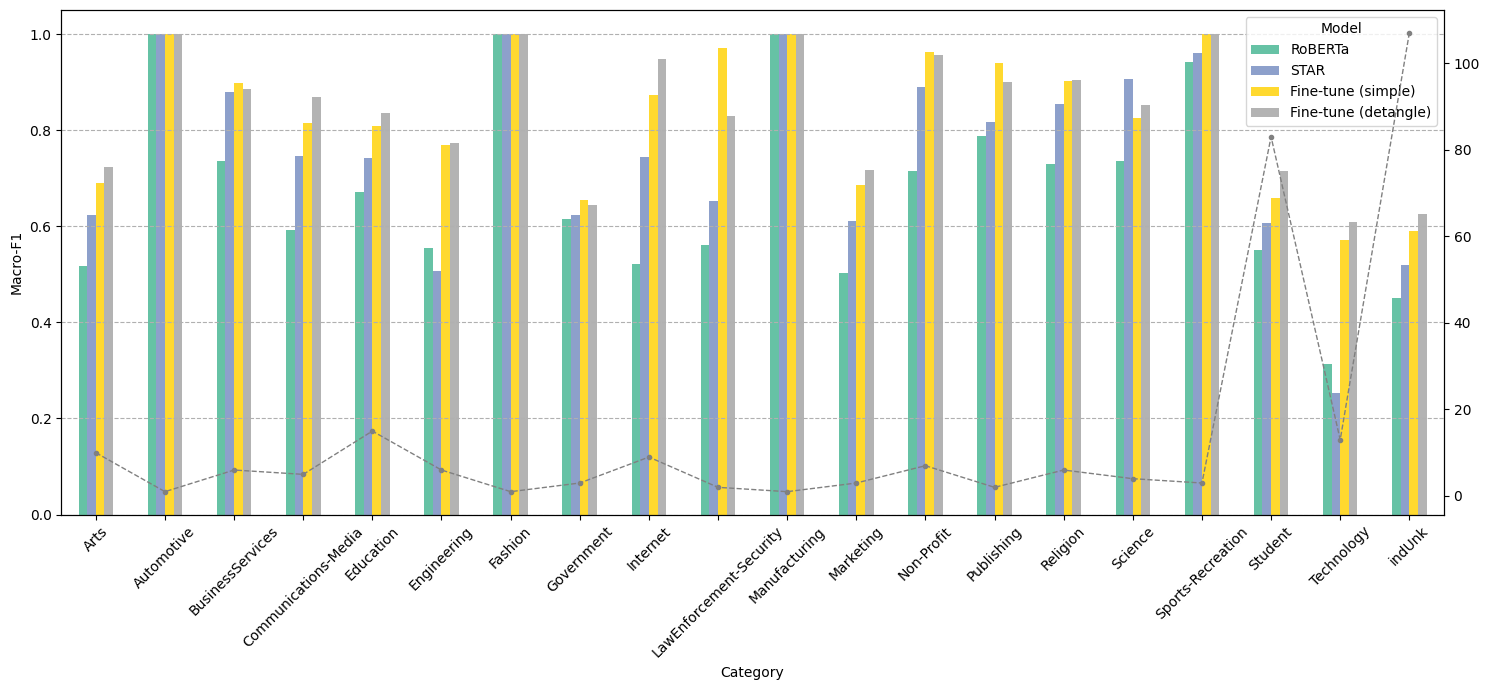}
    \caption{Accuracy on attribution when authors publish across the same category}
    \label{fig:blog-acc}
\end{figure}

In the Fig.~\ref{fig:blog-acc} we split the author set by category, then evaluate. This means it is really easy to confuse authors when topic is exactly the same. The largest relative improvements can be observed when there are more authors in the category, with Students being the most representative of this with 80 elements in the support set. We observe that, if the amount of authors has influence over the accuracy, where smaller sets of authors have less trouble separating style and content, while more populated categories are well attributed by our detangling. 

There are some scenarios where neither fine-tuning nor detangling benefit the model such as government, automotive, security or science where zero-shot STAR or RoBERTa attain better performance (although admittedly not a large difference). There are some edge scenarios where attribution is performed perfectly by all models, which may be due to them having a very low author count (Less than 5). Otherwise, save the aforementioned exceptions, when there is enough support, detanglement shows clear improvement across the board for the examined categories.

Finally, some additional performance metrics are provided in Table~\ref{tab:metrics}. In this table we find the average accuracy attained by each category when performing the global evaluation, paired with the proportion of failures due to confusing authors within the same topic. Our intention is that accuracy rises while category-related errors decrease, and that is exactly what is observed. The detangling method achieves the highest average accuracy per category, while having the lowest ratio of category-related failures. This offers strong indication of the separation achieved by the surrogate model, offering evidence of our findings.

\begin{table}[H]
\centering
\caption{Evaluation of metrics by category and mistakes due to topic}
\label{tab:metrics}
\begin{tabular}{lllll}
\toprule
 & \textbf{RoBERTa} & \textbf{STAR} & \textbf{Fine-tune (simple)} & \textbf{Fine-tune (detangle)} \\
\midrule
Accuracy of each category & 0.3010 & 0.3441 & 0.4192 & \textbf{0.4423} \\
Same topic when miss & 0.1382 & 0.1258 & 0.1193 & \textbf{0.1031} \\
\bottomrule
\end{tabular}
\end{table}

However, the fine-tuned model also increases accuracy per category and reduces topic-related failures, whcih also occurs when going from RoBERTa to STAR. The more style-informed the model is, the less topic-related failures; as such we determine that there is evidence to support that the detangling procedure biases the model strongly towards style rather than content. As we have seen, the procedure is imperfect, prone to failure in some edge categories for the blog dataset.

\subsection{Fanfiction}
The fan-fiction dataset is less challenging in a sense due to the large heterogeneity of topic. Despite this, some interesting evaluations can be performed. First, we show the global results across the entire author set in Table~\ref{tab:fanfic-sum}. As we observed with the blogs, the best model across the board is the detangling procedure. However, the fine-tuning results differ. It is interesting to note that fine tuning the model barely offers any improvement with barely 0.5\% of improvement over zero-shot STAR and less than 0.5\% in terms of F1-score. The dataset is massive and attribution is already decent using STAR, thus simply fine-tuning towards narratives does not have much effect. On the other hand the detangling results in 2\% and 1\% increases over zero-shot, a step above our previous results.

\begin{table}[H]
\centering
\caption{Attribution results across the entire blog author set. Repeated 10 times.}
\label{tab:fanfic-sum}
\begin{tabular}{lrr}
\toprule
 & \textbf{Accuracy} & \textbf{F1 Score} \\
\midrule
\textit{RoBERTa} & 0.7524 & 0.8366 \\
\textit{STAR} & 0.7696 & 0.8426 \\
\textit{Fine-tune (simple)} & 0.7741 & 0.8458 \\
\textit{Fine-tune (detangle)} & \textbf{0.7924} & \textbf{0.8564} \\
\bottomrule
\end{tabular}
\end{table}

Despite the aforementioned increases in performance, can we actually guarantee that these examples are better classified due to the detangling? This is uncertain, thus we observe what happens when we try to apply a model to an extreme edge scenario. We massively reduce the number of authors, but at the same time we pick the authors that publish in the highest number of categories. This heterogeneity leads to failure for non-detangled models as we show in Table~\ref{tab:fanfic-prolifics}. It is simple to observe that the detangling offers a large improvement over both STAR and the fine-tuning. In this scenario we observe that as the number of authors chosen with our criteria goes over 2500, the problem becomes easier.

\begin{table}[H]
\centering

\caption{Evaluation of the top-k most diverse authors.}
\label{tab:fanfic-prolifics}
\begin{tabular}{llrrrrrr}
\toprule
 & Top-K= & 10 & 25 & 100 & 250 & 1000 & 2500 \\
\midrule
\multirow[t]{4}{*}{Accuracy} & RoBERTa & 0.5288 & 0.4493 & 0.3671 & 0.3399 & 0.3598 & 0.3909 \\
 & STAR & 0.7333 & 0.6619 & 0.5108 & 0.4698 & 0.4464 & 0.4526 \\
 & Fine-tune (simple) & 0.7062 & 0.6930 & 0.5065 & 0.4774 & 0.4641 & 0.4697 \\
 & Fine-tune (detangle) & \textbf{0.8158} & \textbf{0.7742} & \textbf{0.5993} & \textbf{0.5473} & \textbf{0.5344} & \textbf{0.5230} \\
\cline{1-8}
\multirow[t]{4}{*}{F1 Score} & RoBERTa & 0.5147 & 0.4470 & 0.3829 & 0.3692 & 0.4193 & 0.4642 \\
 & STAR & 0.7314 & 0.6436 & 0.5069 & 0.4773 & 0.4759 & 0.4996 \\
 & Fine-tune (simple) & 0.7179 & 0.6947 & 0.5067 & 0.4822 & 0.4845 & 0.5076 \\
 & Fine-tune (detangle) & \textbf{0.8194} & \textbf{0.7715} & \textbf{0.5863} & \textbf{0.5481} & \textbf{0.5372} & \textbf{0.5436} \\
\cline{1-8}
\bottomrule
\end{tabular}
\end{table}

In this particular scenario, we find that sometimes zero-shot has better performance that fine-tuning, thus adaption to the target domain is not enough for detanglement. Again, this is strong evidence that our model is performing better on hard attribution problems where topics may be different. The detanglement improvement ranges from 10\% to 7\%  accuracy from zero-shot STAR to the detanglement, extremely large improvements considering our previous results in the global table. In terms of F1 Score, the improvements are also large with similar results between fine-tuned STAR and zero-shot STAR. These observations convey that the model is indeed going beyond category to pick up better stylistic features than before.

\subsection{Challenges}
We test our models on a set of standard challenges for this task. The objective is to observe the loss of accuracy of our detangling fine-tuning with regards to the zero-shot model. If our model has less zero-shot accuracy, it means that the detangling process destroys some amount of the original features. In the Table~\ref{tab:chal-global} we can observe the average metrics obtained by the proposed methods. Interestingly the fan-fiction detanglement is the best model across both metrics, however this could be due to the adaption to the fan-fiction task of the challenge (PAN19). However, this improvement is not present with the fine-tuned STAR on fan-fiction, indicating that there may be something else improving the results with detanglement.

\begin{table}[H]
\centering
\caption{Global metrics for each model across PAN11-19 challenges}
\label{tab:chal-global}
\begin{tabular}{lll}
\toprule
 & & Score  \\
\midrule
\multirow[t]{6}{*}{Accuracy} & RoBERTa & 0.5788 \\
 & STAR & 0.6185 \\
 & Detangled blog & 0.5901 \\
 & Detangled fanfic & \textbf{0.6346} \\
 & Simple blog & 0.5842 \\
 & Simple fanfic & 0.5748 \\
\cline{1-3}
\multirow[t]{6}{*}{macro-F1} & RoBERTa & 0.5327 \\
 & STAR & 0.5693 \\
 & Detangled blog & 0.5339 \\
 & Detangled fanfic & \textbf{0.5795} \\
 & Simple blog & 0.5227 \\
 & Simple fanfic & 0.5224 \\
\cline{1-3}
\bottomrule
\end{tabular}
\end{table}
The detangled blog metrics are worse than STAR in both metrics, however, the simple fine-tuning method is even worse than the detangle. We deduce that the zero-shot capabilities of a fine-tuned model are improved with the detanglement. This also is repeated with the detangled fan-fiction model against the fine-tuned fan-fiction model. The fan-fiction models are similar when they are compared after being fine-tuned, however when the detangled models are compared, the fan-fiction dataset performs much better. We conclude that the detanglement procedure particularly benefits from a larger set of authors.

In Table~\ref{tab:challenge} we finally present the overall performance for all models. Here we observe the outperformance of zero-shot STAR on most problems and scenarios, with sometimes worse performance that the detangled fan-fiction model in some specific problems of PAN11 and PAN19. Most observations performed on these results match our previous results.

\begin{table}[H]
\centering
\tiny
\caption{Metrics on the PAN11-19 challenge evaluation set}
\label{tab:challenge}
\begin{tabular}{lllllllll}
\toprule
 & Challenge & Problem & RoBERTa & STAR & Detangled blogs & Detangled fanfics & Basic blogs & Basic fanfics \\
\midrule
\multirow[t]{18}{*}{Accuracy} & Avg & & 0.5788 & 0.6185 & 0.5901 & \textbf{0.6346} & 0.5842 & 0.5748 \\
\cline{2-9}
 & \multirow[t]{2}{*}{pan11} & problem-large & 0.3777 & \textbf{0.4746} & 0.4538 & 0.4515 & 0.3992 & 0.3700 \\
 &  & problem-small & 0.4545 & 0.5293 & 0.5172 & \textbf{0.5596} & 0.4747 & 0.4707 \\
\cline{2-9}
 & \multirow[t]{6}{*}{pan12} & problem-a & \textbf{1.0000} & 0.8333 & 0.8333 & \textbf{1.0000} & 0.8333 & \textbf{1.0000} \\
 &  & problem-b & \textbf{0.6000} & 0.5000 & 0.5000 & \textbf{0.6000} & 0.5000 & \textbf{0.6000} \\
 &  & problem-c & 1.0000 & 1.0000 & 1.0000 & 1.0000 & 1.0000 & 1.0000 \\
 &  & problem-d & 0.3529 & \textbf{0.4706} & 0.3529 & \textbf{0.4706} & 0.2941 & 0.3529 \\
 &  & problem-i & 0.7143 & 0.7143 & 0.7857 & \textbf{0.8571} & \textbf{0.8571} & 0.7143 \\
 &  & problem-j & 0.6250 & 0.6250 & 0.6250 & 0.6250 & 0.6250 & 0.6250 \\
\cline{2-9}
 & \multirow[t]{4}{*}{pan18} & problem00001 & 0.4937 & \textbf{0.6456} & 0.5823 & 0.6076 & 0.5696 & 0.6203 \\
 &  & problem00002 & 0.5541 & \textbf{0.6486} & 0.5946 & 0.6216 & 0.6081 & \textbf{0.6486} \\
 &  & problem00003 & 0.8000 & \textbf{0.9250} & 0.9000 & 0.8500 & 0.8500 & 0.8500 \\
 &  & problem00004 & 0.5625 & \textbf{0.6250} & \textbf{0.6250} & \textbf{0.6250} & 0.5625 & 0.6875 \\
\cline{2-9}
 & \multirow[t]{5}{*}{pan19} & problem00001 & 0.7540 & 0.7576 & 0.7665 & \textbf{0.7683} & 0.7576 & 0.5294 \\
 &  & problem00002 & 0.3066 & \textbf{0.4453} & 0.3431 & 0.4307 & 0.3504 & 0.3577 \\
 &  & problem00003 & 0.3744 & 0.4076 & 0.3839 & \textbf{0.4645} & 0.4123 & 0.3791 \\
 &  & problem00004 & 0.4799 & \textbf{0.4725} & 0.3626 & 0.4212 & 0.4432 & 0.1868 \\
 &  & problem00005 & 0.3902 & \textbf{0.4394} & 0.4053 & 0.4356 & 0.3939 & 0.3788 \\
\midrule
\multirow[t]{18}{*}{macro-F1} & Avg &  & 0.5327 & 0.5693 & 0.5339 & \textbf{0.5795} & 0.5227 & 0.5224 \\
\cline{2-9}
 & \multirow[t]{2}{*}{pan11} & problem-large & 0.2648 & \textbf{0.3520} & 0.3258 & 0.3311 & 0.3029 & 0.2735 \\
 &  & problem-small & 0.2511 & \textbf{0.3485} & 0.2948 & 0.3260 & 0.2603 & 0.2700 \\
\cline{2-9}
 & \multirow[t]{6}{*}{pan12} & problem-a & \textbf{1.0000} & 0.8222 & 0.8222 & \textbf{1.0000} & 0.8222 & \textbf{1.0000} \\
 &  & problem-b & \textbf{0.5833} & 0.4762 & 0.4762 & 0.5667 & 0.4762 & 0.5667 \\
 &  & problem-c & 1.0000 & 1.0000 & 1.0000 & 1.0000 & 1.0000 & 1.0000 \\
 &  & problem-d & 0.4444 & \textbf{0.6185} & 0.3889 & \textbf{0.6185} & 0.3426 & 0.4519 \\
 &  & problem-i & 0.6429 & 0.6548 & 0.7381 & 0.8095 & \textbf{0.8333} & 0.6667 \\
 &  & problem-j & 0.6000 & 0.5778 & 0.5667 & \textbf{0.6222} & \textbf{0.6222} & 0.6000 \\
\cline{2-9}
 & \multirow[t]{4}{*}{pan18} & problem00001 & 0.5329 & \textbf{0.6225} & 0.5796 & 0.4918 & 0.4645 & 0.4731 \\
 &  & problem00002 & 0.4982 & \textbf{0.6308} & 0.5554 & 0.5244 & 0.5057 & 0.5332 \\
 &  & problem00003 & 0.6957 & \textbf{0.8157} & 0.7623 & 0.6666 & 0.6634 & 0.6184 \\
 &  & problem00004 & 0.5226 & 0.4914 & \textbf{0.5424} & 0.5270 & 0.5127 & 0.5179 \\
\cline{2-9}
 & \multirow[t]{5}{*}{pan19} & problem00001 & 0.6101 & 0.6172 & \textbf{0.6288} & 0.6202 & 0.6251 & 0.5304 \\
 &  & problem00002 & 0.2848 & \textbf{0.4144} & 0.3212 & 0.3922 & 0.3282 & 0.3278 \\
 &  & problem00003 & 0.4189 & 0.4619 & 0.4339 & \textbf{0.5196} & 0.4322 & 0.4201 \\
 &  & problem00004 & 0.3127 & 0.3074 & 0.2201 & \textbf{0.3113} & 0.2700 & 0.2233 \\
 &  & problem00005 & 0.3927 & 0.4664 & 0.4207 & \textbf{0.5243} & 0.4236 & 0.4075 \\
\cline{1-9} \cline{2-9}
\bottomrule
\end{tabular}
\end{table}

\subsection{Discussion}
With these demonstrations, we have proven that the detangling procedure has several properties. First we observe that the detangled version of the model outperforms a contrastive fine-tuning under the same conditions and training time. This exclusively points towards better generalization which could be explained as a regularization property of the semantic embeddings. However, the later experiments on blogs and fan-fiction reveal that the detangled model reduces topic-related failures significantly and effectively matches the same author writing about different topics. This twofold demonstration offers strong evidence that the model obtains higher quality stylistic features by distancing itself from the semantic embedding space.

On the other and we explore our model zero-shot transference to new domains with a battery of standard challenges. Here, we finally see the downsides of detanglement, as the detangled models are frequently worse than the zero-shot STAR. Fine-tuning over-adapts towards a single domain, thus hurting the model performance on other domains. However, there is an exception, where the detangled fan-fiction model can still achieve top performance and compete with the STAR model. This surprising result indicates that a robust enough dataset to detangle results in a model that both, better attributes authors within the adapted dataset, and retains the zero-shot capabilities of the model. This latter conclusion would require further exploration in future works.

Overall, our method shows strong indications that the model has been partially disentangled from the content, obtaining style-oriented embeddings with higher quality than before the procedure. This detanglement also may help retain the generalization to out-of-domain examples as observed in the challenge evaluation.

\section{Conclusions}
\label{sec:conclusions}
In this paper we present and evaluate a method for detangling style from content in authorship attribution tasks. This method is validated on several datasets demonstrating its capabilities at scale. Overall, fine-tuning a model through our method results in better style representations while retaining the generalization abilities of the original zero-shot model. In fact, performing our training on a pre-trained model we observe no technical downsides outside of the time-consuming process of domain adaptation.

The model improves attribution task accuracy on most of the evaluated metrics. Our study reveals that our method achieves better generalization on datasets with a higher author and topic count. The fine-tuning not only improves in-domain metrics of blogs and fan-fiction, but also correctly extends to zero-shot scenarios with barely any performance loss.

Further research is needed to solidify our hypothesis. For example, it would be interesting to observe whether a model could be pre-trained for style using this method in its entirety. It would also be necessary to test if it is possible to achieve this results with another methods of augmentation over the embeddings. Our initial screenings for these techniques were inconclusive and require further examination.

\section*{Acknowledgments}
This work has been supported by the project PCI2022-134990-2 (MARTINI) of the CHISTERA IV Cofund 2021 program; by European Comission under IBERIFIER Plus - Iberian Digital Media Observatory (by the IBERIFIER Plus project, co-funded by the European Commission under the Call DIGITAL-2023-DEPLOY-04, European Digital Media observatory (EDMO) – National and multinational hubs, grant number: IBERIFIER Plus - 101158511.; by EMIF managed by the Calouste Gulbenkian Foundation, in the project MuseAI; by Comunidad Autónoma de Madrid, CIRMA Project (TEC-2024/COM-404); and by TUAI Project (HORIZON-MSCA-2023-DN-01-01, Proposal number: 101168344).

\bibliographystyle{unsrt}  
\bibliography{references}

\begin{thebibliography}{10}

\bibitem{huang2024authorship}
Baixiang Huang, Canyu Chen, and Kai Shu.
\newblock Authorship attribution in the era of llms: Problems, methodologies, and challenges.
\newblock {\em arXiv preprint arXiv:2408.08946}, 2024.

\bibitem{oord2018representation}
Aaron van~den Oord, Yazhe Li, and Oriol Vinyals.
\newblock Representation learning with contrastive predictive coding.
\newblock {\em arXiv preprint arXiv:1807.03748}, 2018.

\bibitem{ouni2023survey}
Sarra Ouni, Fethi Fkih, and Mohamed~Nazih Omri.
\newblock A survey of machine learning-based author profiling from texts analysis in social networks.
\newblock {\em Multimedia Tools and Applications}, 82(24):36653--36686, 2023.

\bibitem{tyoStateArtAuthorship2022}
Jacob Tyo, Bhuwan Dhingra, and Zachary~C. Lipton.
\newblock On the {{State}} of the {{Art}} in {{Authorship Attribution}} and {{Authorship Verification}}, October 2022.

\bibitem{palomino-garibayRandomForestApproach}
Alonso Palomino-Garibay, Adolfo~T Camacho-Gonzalez, Ricardo~A Fierro-Villaneda, Irazu Hernandez-Farias, Davide Buscaldi, Ivan~V Meza-Ruiz, et~al.
\newblock A random forest approach for authorship profiling.
\newblock In {\em Proceedings of CLEF}, 2015.

\bibitem{daneshvarGenderIdentificationTwitter}
Saman Daneshvar and Diana Inkpen.
\newblock Gender identification in twitter using n-grams and lsa.
\newblock In {\em proceedings of the ninth international conference of the CLEF association (CLEF 2018)}. CEUR-WS, 2018.

\bibitem{abbasiWriteprintsStylometricApproach2008}
Ahmed Abbasi and Hsinchun Chen.
\newblock Writeprints: {{A}} stylometric approach to identity-level identification and similarity detection in cyberspace.
\newblock {\em ACM Trans. Inf. Syst.}, 26(2):7:1--7:29, April 2008.

\bibitem{ashrafCrossGenreAuthorProfile}
Shaina Ashraf, Hafiz~Rizwan Iqbal, and Rao Muhammad~Adeel Nawab.
\newblock Cross-genre author profile prediction using stylometry-based approach.
\newblock In {\em CLEF (Working Notes)}, pages 992--999, 2016.

\bibitem{fernquistFourFeatureTypes}
Johan Fernquist.
\newblock A four feature types approach for detecting bot and gender of twitter users.
\newblock In {\em CLEF (Working Notes)}, 2019.

\bibitem{sammutTFIDF2010}
{{TF}}--{{IDF}}.
\newblock In Claude Sammut and Geoffrey~I. Webb, editors, {\em Encyclopedia of {{Machine Learning}}}, pages 986--987. Springer US, Boston, MA, 2010.

\bibitem{sariContinuousNgramRepresentations2017}
Yunita Sari, Andreas Vlachos, and Mark Stevenson.
\newblock Continuous {{N-gram Representations}} for {{Authorship Attribution}}.
\newblock In Mirella Lapata, Phil Blunsom, and Alexander Koller, editors, {\em Proceedings of the 15th {{Conference}} of the {{European Chapter}} of the {{Association}} for {{Computational Linguistics}}: {{Volume}} 2, {{Short Papers}}}, pages 267--273, Valencia, Spain, April 2017. Association for Computational Linguistics.

\bibitem{coyotl-moralesAuthorshipAttributionUsing2006}
Rosa~María Coyotl-Morales, Luis Villaseñor-Pineda, Manuel Montes-y Gómez, and Paolo Rosso.
\newblock Authorship {{Attribution Using Word Sequences}}.
\newblock In José~Francisco Martínez-Trinidad, Jesús~Ariel Carrasco~Ochoa, and Josef Kittler, editors, {\em Progress in {{Pattern Recognition}}, {{Image Analysis}} and {{Applications}}}, pages 844--853. Springer, 2006.

\bibitem{jacoboAuthorshipVerification}
Gianni~X Jacobo, Valeria Dehesa-Corona, Ariel~D Rojas-Reyes, and Helena G{\'o}mez-Adorno.
\newblock Authorship verification machine learning methods for style change detection in texts.
\newblock In {\em CLEF (Working Notes)}, pages 2652--2658, 2023.

\bibitem{muttenthalerAuthorshipAttributionFanFictional}
Lukas Muttenthaler, Gordon Lucas, and Janek Amann.
\newblock Authorship {{Attribution}} in {{Fan-Fictional Texts}} given variable length {{Character}} and {{Word N-Grams}}.
\newblock In {\em CLEF (Working Notes)}, 2019.

\bibitem{hu2024ContrastiveDisentanglement}
Zhiqiang Hu, Thao~Thanh Nguyen, Yujia Hu, Chia-Yu Hung, Ming~Shan Hee, Chun~Wei Seah, and Roy Ka-Wei Lee.
\newblock Contrastive disentanglement for authorship attribution.
\newblock In {\em Companion Proceedings of the ACM Web Conference 2024}, pages 1657--1666, Singapore Singapore, May 2024. ACM.

\bibitem{vaswani2017attention}
Ashish Vaswani, Noam Shazeer, Niki Parmar, Jakob Uszkoreit, Llion Jones, Aidan~N Gomez, {\L}ukasz Kaiser, and Illia Polosukhin.
\newblock Attention is all you need.
\newblock {\em Advances in neural information processing systems}, 30, 2017.

\bibitem{devlin2019bert}
Jacob Devlin, Ming-Wei Chang, Kenton Lee, and Kristina Toutanova.
\newblock Bert: Pre-training of deep bidirectional transformers for language understanding.
\newblock In {\em Proceedings of the 2019 Conference of the North American Chapter of the Association for Computational Linguistics: Human Language Technologies, Volume 1 (Long and Short Papers)}, pages 4171--4186, 2019.

\bibitem{fabienBertAABERTFinetuning}
Ma{\"e}l Fabien, Esa{\'u} Villatoro-Tello, Petr Motlicek, and Shantipriya Parida.
\newblock Bertaa: Bert fine-tuning for authorship attribution.
\newblock In {\em Proceedings of the 17th International Conference on Natural Language Processing (ICON)}, pages 127--137, 2020.

\bibitem{huertas2022part}
Javier Huertas-Tato, Alvaro Huertas-Garcia, Alejandro Martin, and David Camacho.
\newblock Part: Pre-trained authorship representation transformer.
\newblock {\em Human-centric Computing and Information Sciences}, 14, 2024.

\bibitem{liu2019roberta}
Yinhan Liu, Myle Ott, Naman Goyal, Jingfei Du, Mandar Joshi, Danqi Chen, Omer Levy, Mike Lewis, Luke Zettlemoyer, and Veselin Stoyanov.
\newblock Roberta: A robustly optimized bert pretraining approach.
\newblock {\em arXiv preprint arXiv:1907.11692}, 2019.

\bibitem{romanov2019AdversarialDecomposition}
Alexey Romanov, Anna Rumshisky, Anna Rogers, and David Donahue.
\newblock Adversarial decomposition of text representation, April 2019.

\bibitem{goodfellowGenerativeAdversarialNets2014a}
Ian Goodfellow, Jean {Pouget-Abadie}, Mehdi Mirza, Bing Xu, David {Warde-Farley}, Sherjil Ozair, Aaron Courville, and Yoshua Bengio.
\newblock Generative {{Adversarial Nets}}.
\newblock In {\em Advances in {{Neural Information Processing Systems}}}, volume~27. Curran Associates, Inc., 2014.

\bibitem{wegmann2022same}
Anna Wegmann, Marijn Schraagen, and Dong Nguyen.
\newblock Same author or just same topic? towards content-independent style representations.
\newblock {\em arXiv preprint arXiv:2204.04907}, 2022.

\bibitem{sawatphol2022TopicRegularizedAuthorship}
Jitkapat Sawatphol, Nonthakit Chaiwong, Can Udomcharoenchaikit, and Sarana Nutanong.
\newblock Topic-regularized authorship representation learning.
\newblock In Yoav Goldberg, Zornitsa Kozareva, and Yue Zhang, editors, {\em Proceedings of the 2022 Conference on Empirical Methods in Natural Language Processing}, pages 1076--1082, Abu Dhabi, United Arab Emirates, December 2022. Association for Computational Linguistics.

\bibitem{huertas2023understanding}
Javier Huertas-Tato, Alejandro Mart{\'\i}n, and David Camacho.
\newblock Understanding writing style in social media with a supervised contrastively pre-trained transformer.
\newblock {\em Knowledge-Based Systems}, 296:111867, 2024.

\bibitem{li2023angle}
Xianming Li and Jing Li.
\newblock Angle-optimized text embeddings.
\newblock {\em arXiv preprint arXiv:2309.12871}, 2023.

\bibitem{JonathanSchler2023Oct}
Moshe~Koppel Jonathan~Schler.
\newblock {Effects of Age and Gender on Blogging}.
\newblock {\em AAAI}, October 2023.

\bibitem{BibEntry2014Mar}
{Fanfiction.net Re-pack : Free Download, Borrow, and Streaming : Internet Archive}, March 2014.
\newblock [Online; accessed 16. Sep. 2024].

\bibitem{juola2013overview}
Patrick Juola and Efstathios Stamatatos.
\newblock Overview of the author identification task at pan 2013.
\newblock {\em CLEF (Working Notes)}, 1179, 2013.

\bibitem{stamatatos2014overview}
Efstathios Stamatatos, Walter Daelemans, Ben Verhoeven, Martin Potthast, Benno Stein, Patrick Juola, Miguel~A Sanchez-Perez, Alberto Barr{\'o}n-Cede{\~n}o, et~al.
\newblock Overview of the author identification task at pan 2014.
\newblock In {\em CEUR Workshop Proceedings}, volume 1180, pages 877--897. CEUR-WS, 2014.

\bibitem{kestemont2018overview}
Mike Kestemont, Michael Tschuggnall, Efstathios Stamatatos, Walter Daelemans, G{\"u}nther Specht, Benno Stein, and Martin Potthast.
\newblock Overview of the author identification task at pan-2018: cross-domain authorship attribution and style change detection.
\newblock In {\em Working Notes Papers of the CLEF 2018 Evaluation Labs. Avignon, France, September 10-14, 2018/Cappellato, Linda [edit.]; et al.}, pages 1--25, 2018.

\bibitem{kestemont2019overview}
Mike Kestemont, Efstathios Stamatatos, Enrique Manjavacas, Walter Daelemans, Martin Potthast, and Benno Stein.
\newblock Overview of the cross-domain authorship attribution task at $\{$PAN$\}$ 2019.
\newblock In {\em Working Notes of CLEF 2019-Conference and Labs of the Evaluation Forum, Lugano, Switzerland, September 9-12, 2019}, pages 1--15, 2019.

\end{thebibliography}

\end{document}